\documentclass[10pt,twocolumn,letterpaper]{article}

\usepackage{iccv}
\usepackage{times}
\usepackage{epsfig}
\usepackage{graphicx}
\usepackage{amsmath}
\usepackage{amssymb}
\usepackage{multirow}
\usepackage{amsthm}
\usepackage{bbm}
\usepackage{float}
\usepackage{booktabs}
\usepackage[ruled,vlined]{algorithm2e}
\usepackage{setspace}
\usepackage{bm}

\usepackage{xcolor}

\newcommand{\set}[1]{\mathcal{#1}}

\usepackage[pagebackref=true,breaklinks=true,letterpaper=true,colorlinks,bookmarks=false]{hyperref}

\iccvfinalcopy 

\def\iccvPaperID{8725} 

\ificcvfinal\fi

\begin{document}

\title{STFAR: Improving Object Detection Robustness at Test-Time by Self-Training with Feature Alignment Regularization}

\author{Yijin Chen\textsuperscript{1},
Xun Xu\textsuperscript{2\textdagger}, Yongyi Su\textsuperscript{1},
and Kui Jia\textsuperscript{1\textdagger}\\
\textsuperscript{1}South China University of Technology,\quad \textsuperscript{2}I2R, A-STAR\\ 
}

\maketitle

\let\thefootnote\relax\footnote{\textsuperscript{\textdagger}Correspondence to Xun Xu $<$alex.xun.xu@gmail.com$>$ \& Kui Jia $<$kuijia@scut.edu.cn$>$.}

\ificcvfinal\thispagestyle{empty}\fi

\begin{abstract}
Domain adaptation helps generalizing object detection models to target domain data with distribution shift. It is often achieved by adapting with access to the whole target domain data. In a more realistic scenario, target distribution is often unpredictable until inference stage. This motivates us to explore adapting an object detection model at test-time, a.k.a. test-time adaptation~(TTA). In this work, we approach test-time adaptive object detection~(TTAOD) from two perspective. First, we adopt a self-training paradigm to generate pseudo labeled objects with an exponential moving average model. The pseudo labels are further used to supervise adapting source domain model. As self-training is prone to incorrect pseudo labels, we further incorporate aligning feature distributions at two output levels as regularizations to self-training. To validate the performance on TTAOD, we create benchmarks based on three standard object detection datasets and adapt generic TTA methods to object detection task. Extensive evaluations suggest our proposed method sets the state-of-the-art on test-time adaptive object detection task.

\end{abstract}

\section{Introduction}

Object detection is a fundamental task in computer vision research and has enabled numerous applications including autonomous driving, robotics, etc. With the advent of deep neural networks, we have observed unprecedented performance of object detection on different types of natural images~\cite{girshick2015fast,ren2015faster,carion2020end}. Despite the encouraging progress on developing more efficient and accurate object detection algorithms, the robustness of these algorithms are often overlooked. Recent studies have revealed that by injecting photorealistic corruptions to natural images, the accuracy of existing object detection algorithms will suffer greatly~\cite{michaelis2019dragon}. 

\begin{figure}
    \centering
    \includegraphics[width=\linewidth]{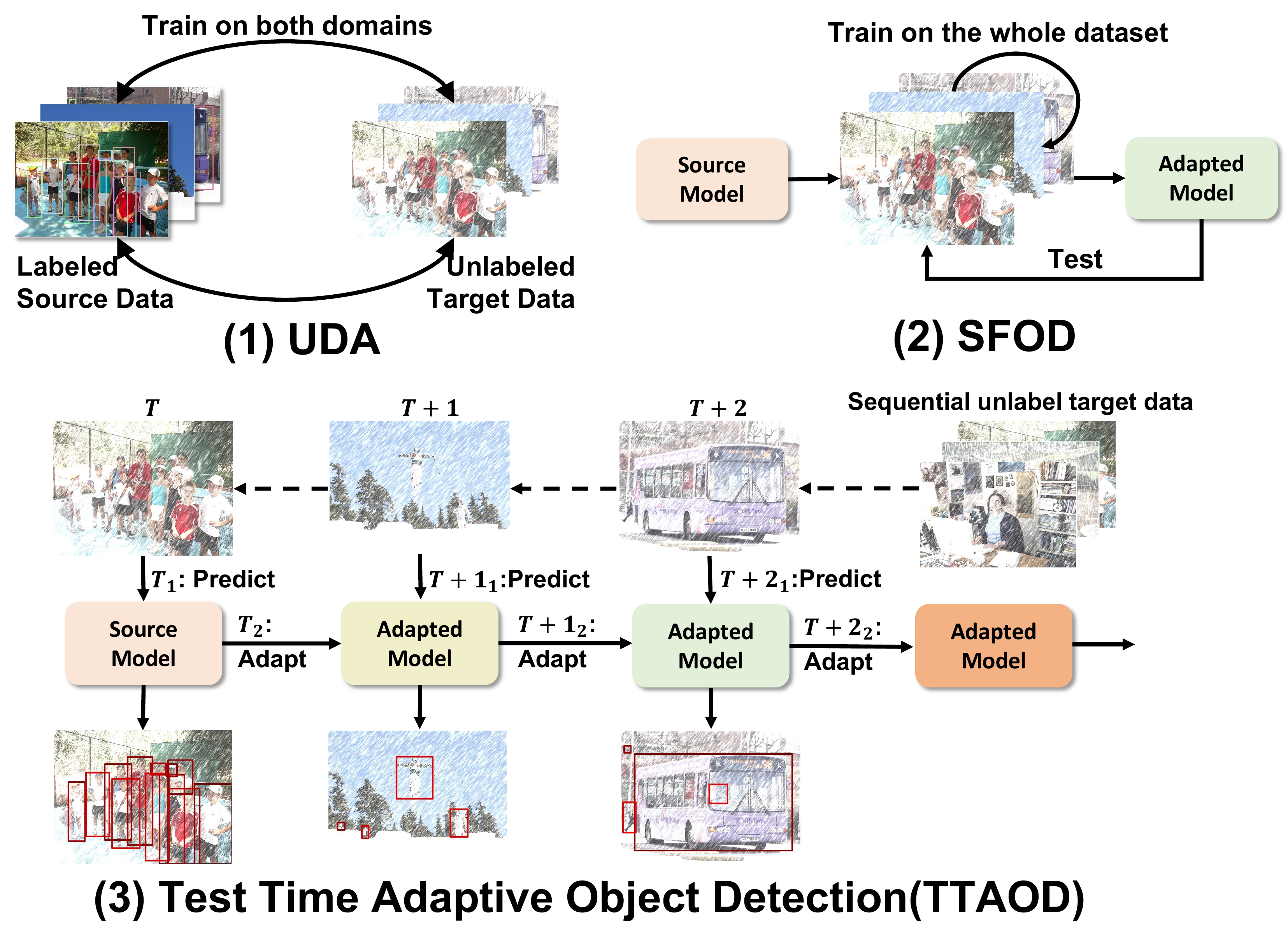}
    \caption{We illustrate the difference between the adopted test-time adaptation protocol and existing UDA and SFOD protocols. UDA requires access to both source and target domain data for adaptation. SFOD requires access to all target domain testing data for adaptation. In contrast, TTA sequentially adapts to target domain testing data on-the-fly.}
    \label{fig:protocol}
\end{figure}

To remedy the robustness of object detection models, unsupervised domain adaptation approaches are employed to learn domain invariant features to improve model generalization~\cite{ganin2015unsupervised}. UDA assumes both source and target domain data samples are available during training a domain generalizable model. This assumption, however, is only applicable to the scenarios where source domain data is accessible and target domain distribution is static. Unfortunately, in a more realistic scenario, source domain data may not be available for adaptation due to privacy issues~\cite{li2020model}. Hence, simultaneously training invariant representation on both source and target domain data is prohibited. Alternative to the strong assumptions made in UDA, source-free domain adaptation~(SFDA)~\cite{liang2020we} relaxes the access to source domain data for domain adaptation. Extension to object detection, namely source-free object detection~(SFOD), has been attempted by self-training with pseudo label~\cite{li2021free} or style transfer~\cite{li2022source}. Although SFDA advances further towards a more realistic domain adaptation setup, we argue that there are still realistic challenges remaining unresolved by SFDA. First, target domain distribution, e.g. the types of corruptions, are often unpredictable before testing begins. For example, it is unrealistic to assume the specific corruption that could happen subject to changing weather or lighting conditions on a new camera, until the testing samples are observed. SFDA will thus struggle to adapt to a testing distribution totally unknown before testing starts. Moreover, testing samples arrive in a sequential manner, predictions on testing samples should be made instantly upon the arrival of a new testing sample~\cite{su2022revisiting}. Since SFDA requires access to the whole target domain samples for adaptation, it fails to enable simultaneous inference and adaptation on-the-fly.

In response to the unpredictability of target domain distribution and demand for simultaneous inference and adaptation, test-time adaptation~(TTA)~\cite{sun2020test, wang2020tent,su2022revisiting} emerged as a solution to adapt model weights to target distribution on-the-fly. TTA advocates a protocol that adaptation is carried out sequentially at test-time and predictions are made instantly~\cite{su2022revisiting}, an illustration of the difference between UDA, SFOD and TTA object detection is presented in Fig.~\ref{fig:protocol}. It is often achieved by dynamically aligning source and target distributions~\cite{su2022revisiting}, self-training with pseudo labels~\cite{chen2022contrastive} or introducing self-supervised tasks~\cite{sun2020test,liu2021ttt++}. However, the existing TTA approaches are almost exclusively developed for image classification tasks~\cite{sun2020test,wang2020tent,liu2021ttt++,chen2022contrastive,su2022revisiting}. It still remains elusive of how to adapt TTA methods to object detection tasks. 


In this work, we approach test-time adaptive object detection~(TTAOD) from two perspectives. First of all, as self-training~(ST) has demonstrated great success in semi-supervised learning~\cite{sohn2020fixmatch,xu2021end} and domain adaptation~\cite{liu2021cycle} by exploiting unlabeled data, we propose to employ self-training for TTAOD by learning from the unlabeled testing samples. This is often achieved by first predicting pseudo labelled objects on the testing sample which are then used for supervising the network training. However, as we empirically revealed in Fig.~\ref{fig:cumulative}, the performance of applying ST alone may gradually degrade upon seeing more unlabeled data. This is probably caused by learning from a accumulation of incorrect pseudo labels, which is also referred to as confirmation bias~\cite{arazo2020pseudo}. Therefore, additional regularization is required to stabilize self-training for TTAOD.
Alternative to self-training, distribution alignment has demonstrated success in test-time adaptation~\cite{liu2021ttt++,su2022revisiting}.
Hence, we introduce distribution alignment as a regularization to self-training for TTAOD.
Specifically, we first propose to align the backbone feature distribution between source and target domains, which is referred to as the \textbf{global feature alignment}. By doing so, target domain backbone feature will be better aligned with source distribution thus easing the difficulty of reusing the downstream RPN and predictor networks. In contrast to TTA approaches for classification, we further notice that a generic object detection predictor involves a classifier for predicting semantic labels and regressor for predicting spatial location. Therefore, reusing source domain classifier and regressor, a commonly adopted practice in TTA, requires eliminating covariate shift at the foreground features. For this purpose, we further propose to align distributions at ROI feature map level, which is referred to as the \textbf{foreground feature alignment}.

To validate the effectiveness of the proposed method, we establish a benchmark for test-time adaptive object detection task. We created corrupted target domain data from three standard object detection datasets and adapted state-of-the-art TTA methods for object detection task. Extensive experiments are carried out on these datasets. The contributions of this work are summarized as below:

\begin{itemize}
    \item We aim to improve the robustness of object detection algorithms to corruptions that are not predictable before testing. The model must be adapted to testing data distribution at test-time, which is referred to as test-time adaptive object detection~(TTAOD). 
    \item Test-time adaptive object detection is enabled by self-training~(ST) on the testing data. For more stable ST we introduce source and target domain feature alignment at both global and foreground level as regularization. The combined model enables more stable and effective TTA performance.
    \item We adapted existing TTA methods to object detection tasks and created a benchmark for test-time adaptive object detection task. Evaluations on three object detection datasets demonstrated the effectiveness of the proposed method.
\end{itemize}
\section{Related Works}






\begin{figure*}[!ht]
\centering
\includegraphics[width=0.85\textwidth]{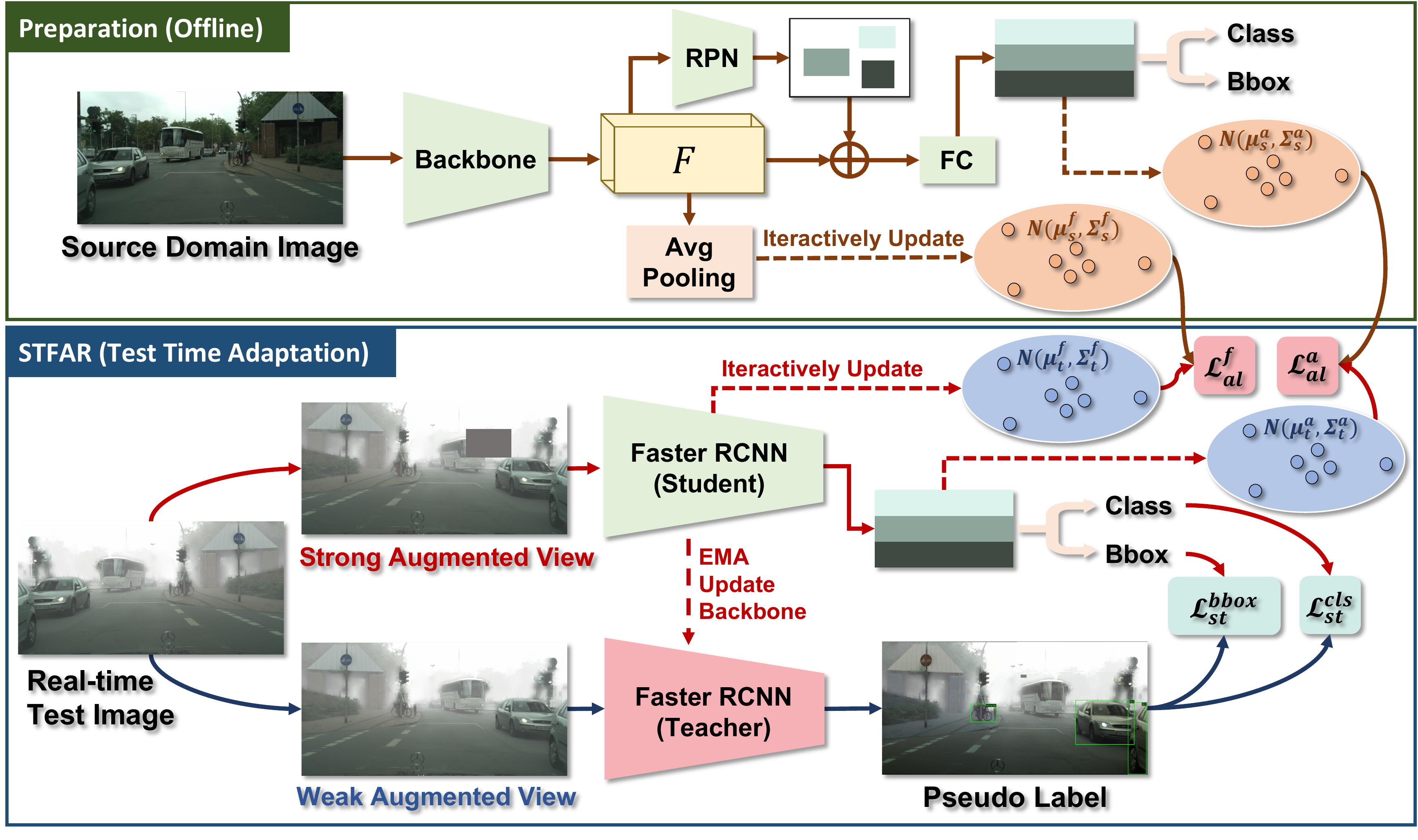}
\caption{The overview of the proposed method, STFAR. In the source domain, STFAR computes the feature distributions at both global and foreground level in an offline manner. During test-time adaptation, self-training is applied by predicting pseudo labels with teacher network. The student network is then supervised by the pseudo labels. Self-training is further regularized by distribution alignment for improvement robustness.}
\vspace{-0.0cm}
\label{fig:overview}
\end{figure*}


\subsection{Domain Adaptive Object Detection}
In the recent years, several Unsupervised Domain Adaptive Object Detection~(UDAOD) studies~\cite{chen2018, Saito_2019_CVPR, Chen_2020_CVPR, He_2019_ICCV, Zhao2020, Xu_2020_CVPR, Khodabandeh_2019_ICCV, li2021, Kim_2019_CVPR, wang_2021_tip, Xie_2019_ICCV, Kim_2019_ICCV, He_2022_CVPR} have been proposed to alleviate the impact from the domain gap cross domains in object detection task. These methods can be roughly divided into the following categories.
i) Aligning the distributions of source and target domain in different layers and levels, e.g., DA-Faster~\cite{chen2018}, a pioneer in UDAOD, proposes a domain adaptive Faster R-CNN to reduce the domain discrepancy on both image and instance levels by adopting two different level domain classifiers and employing the adversarial training strategy. SWDA~\cite{Saito_2019_CVPR} proposes to align local features on shallow layers and image-level features by Focal Loss~\cite{Lin_2017_ICCV} on deep layers, i.e. strong local and weak global alignments. Similar to SWDA, Dense-DA~\cite{Xie_2019_ICCV}, MAF~\cite{He_2019_ICCV}, HTCN~\cite{Chen_2020_CVPR} and SSA-DA~\cite{Zhao2020} align features on multiple layers by adversarial training. ICR-CCR~\cite{Xu_2020_CVPR} leverages the categorical consistency between image-level and instance-level predictions to re-weight the instance-level alignment.
ii) Training from noisy labels, a.k.a. self-training, e.g., NL~\cite{Khodabandeh_2019_ICCV} and WST-BSR~\cite{Kim_2019_ICCV}. 
iii) Based on sample generation strategy. In this line of works, DD-MRL~\cite{Kim_2019_CVPR} leverages an image-to-image translation via GAN to generate various distinctive shifted domains from the source domain. AFAN~\cite{wang_2021_tip} obtains the intermediate domain (fusing the source and target domains) by interpolation. UMT~\cite{Xie_2019_ICCV} and TDD~\cite{He_2022_CVPR} utilize both the source-like and target-like images to perform the cross-domain distillation. To enhance the robustness of the cross-domain model, both UMT and TDD utilize the teacher-student learning scheme, in which UMT adopts Mean Teacher~\cite{NIPS2017_68053af2} and TDD adopts Dual-branch detection network. In this work, a momentum-updated Faster R-CNN is performed for more stability in test-time adaptation. 
Although the excellent performance is reached, all UDAOD methods require access to the source domain data during the adaptation process. When the source data is not accessible due to privacy issues or storage overhead, more challenging settings are emerged with source-free domain adaptation~\cite{li2020model, liang2020we, li2022source} and test-time adaptation~\cite{sun2020test, wang2020tent}.

\subsection{Source-Free Object Detection}
Without access to the source data, Source-Free Domain Adaptation~(SFDA) aims to explore how to rely only on unlabeled target data to adapt the source pre-trained model to the target domain. In the classification task, 3C-GAN~\cite{li2020model} generate labeled target-like samples through conditional GAN~\cite{conditionalGAN} for training. SHOT~\cite{liang2020we} generates pseudo labels for each target samples and performs self-training process and information maximization to ensure class balanced. 
Recently, a few SFDA methods are used for alleviating the domain gap in object detection task when source data is not accessible, called Source-Free Object Detection~(SFOD). SED~\cite{li2021free} proposes self-entropy descent policy to search a confident threshold for pseudo labels generation. HCL~\cite{NEURIPS2021_1dba5eed} proposes historical contrastive instance discrimination to encourage the consistency between current representation and historical representations. LODS~\cite{li2022source} enhances the style of the target image via style enhancement module and reduce the style degree difference between the original image and the enhanced one. It's not long ago that A$^2$SFOD~\cite{Chu2023adver} is proposed to split target data into source similar part and dissimilar part and align them by adversarial training.
It has been demonstrated SFOD mthods performs well on cross-domain object detection even compared against UDAOD methods~\cite{li2022source}. Nevertheless, SFOD requires adaptation performed in target domain for multiple epochs. In a more realistic DA scenario where inference and adaptation must be implemented simultaneously, in the other words, real-time target domain data cannot be collected in bulk in advance, SFOD will no longer be effective.

\subsection{Test-Time Adaptation}
Collecting target domain samples in bulk in advance and transferring source model in an offline manner restricts the application to adapting to a static known target domain. To allow fast and online adaptation on unlabeled target domain, Test-Time Adaptation~(TTA)~\cite{sun2020test, wang2020tent} emerges. TTT-R~\cite{sun2020test}, a pioneer to this line, adapt the model on-the-fly by an auxiliary self-supervised task. Tent~\cite{wang2020tent} first proposes a fully test-time adaptation method without any auxiliary branch.  Following TTT-R and Tent, many effective methods, e.g., aligning source and target distribution~\cite{liu2021ttt++, su2022revisiting}, self-training with pseudo labels~\cite{chen2022contrastive}, test-time normalization~\cite{lim2023ttn}, anti-forgetting test-time adaptation~\cite{pmlr-v162-niu22a}, prototype-learning~\cite{iwasawa2021test}, more realistic test-time training/adaptation~\cite{su2022revisiting, su2023revisiting, niu2023towards} and etc., have been proposed. However, the existing TTA methods are almost specific to image classification task rather than object detection task, which brings more challenges to the adaptation on-the-fly. In this work, we are devoted to study a more realistic and practical problem on adapting to real-time target domain for object detection on-the-fly, and denoted this setting as Test-Time Adaptive Object Detection~(TTAOD).



\section{Methodology}

In this section, we first provide an overview of test-time adaptation protocol. Then, we introduce self-training for object detection and how feature alignment could regularize self-training fore more resilient test-time training.

\subsection{Overview of Test-Time Adaptation}

Test-time training aims to adapt model weights to target domain distribution in parallel to the inference. We denote the source domain labeled data as $\set{D}_s=\{x_i,y_i\}$ where $y_i=\{\set{B}_i,\set{C}_i\}$ are the ground-truth box annotation and class labels respectively. We further denote the backbone network as $f_i=f(x_i;\Theta)\in\mathbbm{R}^{H\times W\times C}$, proposals after RPN and ROI pooling as $a_i\in\mathbbm{R}^{N_a\times D}$ and the predictors as $h_c(a_i)$ for semantic label classification and $h_r(a_i)$ for location and size regression. An object detection model is trained on the source domain labeled data by optimizing the classification and regressions losses. When the model is deployed for testing on the target domain unlabeled data $\set{D}_t=\{x_j\}$, we assume the testing samples are sequentially streamed and predicted by the model first, and the model weights $\Theta$ are then updated after observing a batch of testing samples. In the following sections, we shall elaborate the details of achieving test-time adaptation for object detection by self-training with distribution alignment regularization. 

\subsection{Test-Time Adaptation by Self-Training}

Self-training~(ST) has demonstrated tremendous effectiveness for semi-supervised learning~\cite{sohn2020fixmatch}. ST often predicts pseudo labels on the unlabeled data samples and the most confident pseudo labels are used for supervising model training. 
In this work, we adopt an approach similar to semi-supervised learning~\cite{tarvainen2017mean}, two networks are maintained throughout the training stage, namely the student network $f(x;\Theta)$ and the teacher network $f(x;\hat{\Theta})$. The teacher network weights are the exponential moving average of student ones as below.

\begin{equation}
    \hat{\Theta} = \beta\hat{\Theta} + (1-\beta)\Theta
\end{equation}

An input image is first applied with one strong augmentation $\mathcal{S}(x)$ and one weak augmentation $\mathcal{W}(x)$. We adopt the augmentation strategy proposed in \cite{xu2021end} as strong augmentation. 
 The teacher model will predict objects on the weak augmented sample $\mathcal{W}(x)$ and obtain a number of pseudo labeled objects $\set{P}=\{\hat{b}_i,\hat{y}_i\}$ where $\hat{b}_i$ and $\hat{y}_i$ refer to as bounding box coordinate and object class label. The student model will treat the pseudo labeled objects as ground-truth and standard supervised learning losses apply to the student model. Specifically, we optimize the classification $\mathcal{L}^{cls}_{st}$ and regression $\mathcal{L}_{st}^{reg}$ losses on the student model branch, where the classification and regression losses follow the definitions in \cite{ren2015faster}.



\subsection{Test-Time Adaptation by Distribution Alignment}

Self-training~(ST) alone is prone to the influence of incorrect pseudo labels, a.k.a. confirmation bias~\cite{arazo2020pseudo}. The situation is less severe in semi-supervised learning as the labeled loss serves as a strong regularization to self-training. As no labeled data exists in test-time adaptation, direct ST without regularization is exposed to the risk of failing on unlabeled testing data.
As empirically revealed in Sect.~\ref{sect:further}, the performance of ST may degrade after certain training iterations. Therefore, to improve the robustness of self-training we further incorporate distribution alignment, which has demonstrated success for test-time adaptation/training~\cite{liu2021ttt++,su2022revisiting}, as regularization for self-training.

Existing distribution alignment is developed for classification task. For object detection task, we propose to align two types of features, the backbone feature and foreground feature. Aligning distribution at these two features allows us to achieve better reusing of RPN network and box predictors. Specifically, we use multi-variate Gaussian distributions in the source domain {as $N(\mu_s^f, \Sigma_s^f)$ and $N(\mu_s^a, \Sigma_s^a)$} to characterize global and foreground feature distribution. In a typical faster RCNN framework, the backbone network outputs global feature map $f_i\in\mathbbm{R}^{C\times H\times W}$ and proposal features after RPN and ROI pooling $a_i\in\mathbbm{R}^{N_p\times D}$. To obtain a single vectorized backbone feature and foreground feature for each individual image $x_i$ for estimating the distribution, we first do average pooling over global and foreground features respectively as below.

\begin{equation}\label{eq:g_feature}
\begin{split}
g^f(x_i)=\frac{1}{HW}\sum\limits_{h,w} z_{ihw};\;
g^a(x_i)=\frac{1}{N_a}\sum\limits_{j=1\cdots N_a} a_{ij}
\end{split}
\end{equation}

With vectorized features for each image, we estimate the distribution information by Eq.~\ref{eq:gauss}, the same estimation applies to foreground features, denoted as $\mu_s^a,\;\Sigma_s^a$. 

\begin{equation}\label{eq:gauss}
\begin{split}
    &\mu_s^f = \frac{1}{|\set{D}_s|}\sum\limits_{x_i\in\set{D}_s} g^f(x_i),\\
    &\Sigma_s^f = \frac{1}{|\set{D}_s|}\sum\limits_{x_i\in\set{D}_s}(g^f(x_i)-\mu_s^f)(g^f(x_i)-\mu_s^f)^\top
\end{split}
\end{equation}


\noindent\textbf{Distribution Alignment}: Aligning target distribution to the source domain is achieved by minimizing the symmetric KL-Divergence between two multi-variate Gaussian distributions as in Eq.~\ref{eq:alignment}. As KL-Divergence between two Gaussian distributions has a closed-form solution, we can directly use gradient descent methods to optimize the distribution alignment objective. Alignment between the foreground feature distributions follows the same definition by substituting $\mu^f$ and $\Sigma^f$ with $\mu^a$ and $\Sigma^a$, resulting in $\mathcal{L}_{al}^a$.

\begin{equation}\label{eq:alignment}
\begin{split}
    \mathcal{L}_{al}^f=&D_{KL}(\mathcal{N}(\mu_s^f,\Sigma_s^f)||\mathcal{N}(\mu_t^f,\Sigma_t^f)) \\
    &+ D_{KL}(\mathcal{N}(\mu_t^f,\Sigma_t^f)||\mathcal{N}(\mu_s^f,\Sigma_s^f))
\end{split}
\end{equation}

\noindent\textbf{Incremental Target Domain Update}: Although the source domain distributions can be updated in an off-line manner on all available source-domain training samples, it is not equally trivial to estimate the distribution in the target domain under a TTT protocol. Estimating the distribution within a single minibatch of testing samples as the target domain distribution is subject to randomness of testing data distribution within a small temporal window, e.g. testing samples are not drawn i.i.d. from the target domain distribution~\cite{gongnote2022}. Therefore, we propose to estimate the true target domain distribution in an exponential moving average manner. In specific, we use a hyperparameter $\gamma$ to control the contribution of current minibatch and the target domain distribution can be derived incrementally following the rules in Eq.~\ref{eq:update}, where $\set{B}$ indicates a minibatch of testing samples.

\begin{equation}\label{eq:update}
\resizebox{0.9\linewidth}{!}{
$
\begin{split}
    &\mu_t =  \mu_t + \delta\\
    &\Sigma_t = \Sigma_t +   \gamma\sum\limits_{x_i\in\set{B}}[(g(x_i)-\mu_t)(g(x_i)-\mu_t)^\top-\Sigma_t] - \delta\delta^\top\\
    &\delta = \gamma\sum\limits_{x_i\in\set{B}}(g(x_i)-\mu_t)
\end{split}
$
}
\end{equation}

\subsection{TTA for Object Detection Algorithm}

In this section, we summarize the overall algorithm on test-time adaptation for object detection. On the source domain, we summarize the backbone features and foreground features with two Gaussian distributions in an offline manner. During test-time adaptation, we make object detection prediction for each testing sample for instant inference and simultaneously update the distribution estimations in the target domain. When a minibatch of testing samples are accumulated, we update the model weights through gradient descent. A detailed description of test-time adaptation algorithm is summarized in Alg.~\ref{alg:main}.

\begin{algorithm}
\setstretch{1.2}
\caption{Test-time adaptive object detection algorithm.}\label{alg:main}

\textbf{Input:} Testing sample batch $\set{B}^t=\{x_i\}_{i=1 \cdots N_B}$.

\textcolor{gray}{\# Inference Stage:}\\
\For{$x_i \leftarrow 1$ \KwTo $N_B$}{
    Predict objects: $h_c(a(f(x_i)), h_r(a(f(x_i))$
}

\textcolor{gray}{\# Adaptation Stage:}\\
\For{$x_i \leftarrow 1$ \KwTo $N_B$}{
    Augmentation: $\mathcal{W}(x_i), \mathcal{S}(x_i)$\;
    Pseudo label prediction: $\set{P}=\{\hat{b}_{ij}, \hat{y}_{ij}\}$\;
    Incremental distribution update: Update $\mathcal{N}(\mu_t^f, \Sigma_t^f)$ and $\mathcal{N}(\mu_t^a, \Sigma_t^a)$ by Eq.~\ref{eq:update}\;
    Test-time adaptation loss: $\mathcal{L}_{tta} = \lambda_{st}^{cls}\mathcal{L}_{st}^{cls} + \lambda_{st}^{reg}\mathcal{L}_{st}^{reg} + \lambda_{al}^f\mathcal{L}_{al}^f + \lambda_{al}^a\mathcal{L}_{al}^a$\;
    Gradient descent update: $\Theta = \Theta - \alpha \nabla \mathcal{L}_{tta}$
}
\end{algorithm}
\vspace{-0.3cm}

\section{Experiments}

In this section, we validate the effusiveness of STFAR on test-time adaptive object detection task. We adopt the corrupted versions of 3 standard object detection datasets to create a test-time adaptation benchmark for object detection. Existing test-time adaptation methods are adapted to object detection task for comparison. 

\begin{figure*}[!ht]
\centering
\includegraphics[width=0.99\textwidth]{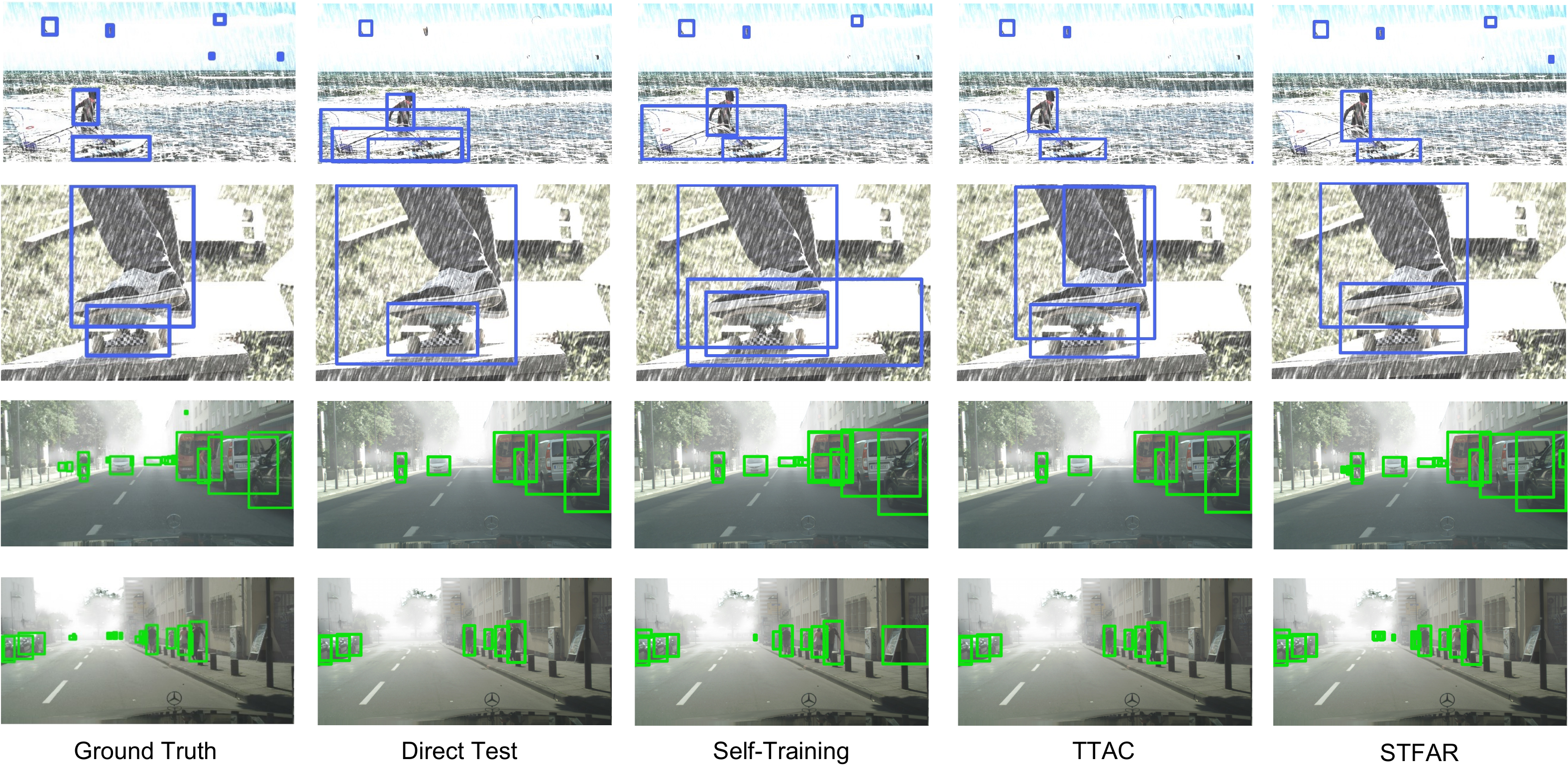}
    \caption{Illustration of the detection results on the target domain, the first two lines and the last two lines represent the scenarios of \textbf{ COCO} → \textbf{ COCO-C}(Snow corruption) and \textbf{Cityscapes} → \textbf{Foggy-Cityscapes}.}
\label{fig:qualitative}
\vspace{-0.4cm}
\end{figure*}

\subsection{Dataset and Evaluation Protocol}
We evaluate on three standard object detection datasets.
\textbf{MS-COCO}~\cite{lin2014microsoft} provided two training data sets,where train2017 contained 118k labeled images and unlabed2017 contained 123k unlabeled images. In addition,  val2017 datsset containing 5k images is provided for verification. We create a corrupted version on  COCO, termed  COCO-C by employing an image corruption package~\cite{michaelis2019dragon}, which consists of 15 types of corruptions. For TTA experiments on  COCO-C, we use the standard  COCO training set to pretrain a source model, target domain is created by applying the corruptions to the validation set of  COCO. 
\textbf{Pascal}~\cite{everingham2015pascal} contains 20 categories of natural images. We employ approximately 15K images from the training and validation sets of the PASCAL VOC 2007 and 2012 to pre-train the source model. We follow a similar way to  COCO-C to generate PASCAL-C on the test dataset of PASCAL2007 containing 4952 images as the target domain.
\textbf{Cityscapes}~\cite{cordts2016cityscapes} consists of 2,975 training images and 500 testing images with 8 categories of objects. The \textbf{Foggy-Cityscapes}~\cite{sakaridis2018semantic} dataset was created by simulating images captured under foggy weather condition. Three levels of corruptions are generated in Foggy-Cityscapes for each image, controlled by a hyper-parameter $\beta=0.005, 0.01,  0.02$, In this work, we choose the most difficult corruption level $\beta=0.02$ for evaluation. %

\subsection{Implementation Details}
\noindent\textbf{Hyperparameters}: We evaluate with ResNet-50 and ResNet-101~\cite{he2016deep} as backbone network following the Faster RCNN object detection framework~\cite{ren2015faster}. We optimize the backbone network by SGD with momentum on the three datasets. On COCO-C, we set batchsize to 8 and learning rate to 1e-4, on Pascal-C we set batchsize to 8 and learning rate to 1e-5, and on Foggy-CityScapes we set batchsize to 4 and learning rate to 1e-5. For other hyperparameters, we set $\lambda_{st}^{cls}=\lambda_{st}^{reg}$ to 1.0 on three datasets. Specially, on COCO-C dataset we set $\lambda_{al}^f$ to 0.1, $\lambda_{al}^a$ to 0.01 and $\gamma$ to $\frac{1}{64}$, on Foggy-Cityscapes dataset we set $\lambda_{al}^f$ to 0.1, $\lambda_{al}^a$ to 1.0 and $\gamma$ to $\frac{1}{256}$, and on Pascal-C dataset, we set $\lambda_{al}^{f}$ to 2.0, $\lambda_{al}^a$ to 0.1 and $\gamma$ to $\frac{1}{128}$.

\noindent\textbf{Data Augmentation}: The strong augmentation consists of common augmentations, including scale jitter, solarize jitter, brightness jitter, contrast jitter, sharpness jitter, translation, rotation and translation. In addition, the strong augmentation also includes RandErase, which randomly samples a few patches (less than 5) at random locations and erases its pixels with a fix valued, simulating  occlusions. The weak augmentation only contains random resize and random flipping, similar to the test augmentation. Under the weak augmentation, teacher model can provide more precious pseudo label to instruct the student model.

\noindent\textbf{Competing Methods}: We adapt the following generic state-the-of-art test-time adaptation methods to object detection task. Direct testing (\textbf{Direct Test}) without adaptation simply does inference on target domain with source domain model. Test-time normalization (\textbf{BN})~\cite{ioffe2015batch} moving average updates the batch normalization statistics in the backbone network by the testing data. Test-time entropy minimization (\textbf{TENT})~\cite{wang2020tent} updates the parameters of all batch normalization layers in the backbone network by
minimizing the entropy of the model predictions on the testing data. Test-time classifier adjustment (\textbf{T3A}) \cite{iwasawa2021test}
computes target prototype representation for each category using testing data and make predictions
with updated prototypes. We allow T3A to update the classification predictor only. Source Hypothesis Transfer
(\textbf{SHOT}) \cite{liang2020we} freezes the linear classification head and trains the target-specific feature extraction module by exploiting balanced category assumption and self-supervised pseudo-labeling in the target domain. 
\textbf{Self-Training}~\cite{xu2021end} was originally developed semi-supervised object detection. We adapt it to TTA by only exploiting the unsupervised learning component.
\textbf{TTAC}~\cite{su2022revisiting} discovers clusters in both source and target domain and match the target clusters to the source ones to improve generalization.
Finally, we present our own approach \textbf{STFAR} by doing self-training with distribution alignment regularization. STFAR,  SHOT and TTAC update the backbone weights during test-time adaptation.

\subsection{Test-Time Adaptation for Corrupted Target Domain}

We evaluate test-time adaptation~(TTA) performance on  COCO-C with results presented in Tab.~\ref{tab: COCO}. When target domain is contaminated with corruptions, the mAP drops substantially from $44.6\%\rightarrow15.9\%$ with ResNet50 as backbone, indicating the corruptions in target domain poses great challenge to the generalization of source domain model. When BN, TENT and T3A are adapted to TTA for object detection, we observe a further drop of performance. It suggests only updating the batchnorm components or classifier weights is not enough for tackling the distribution shift caused by corruptions. SHOT and Self-Training demonstrate improved results from direct testing (Direct Test), suggesting self supervising with pseudo label provides a viable way to TTA for object detection. TTAC serves as another strong baseline for TTA. This indicates distribution alignment is a very effective way to adapt source domain model with corruption as distribution shift. Finally, our STFAR combines self-training with distribution alignment and achieves the best results. We further report TTA performance on PASCAL-C dataset in Tab.~\ref{tab:pascal} and Foggy-CityScapes in Tab.~\ref{tab:cityscapes}, respectively. We draw similar conclusions from these results. STFAR consistently outperforms the other competing methods and improves from the baseline by a large margin. 

\noindent\textbf{Qualitative Results}: We visualize object detection results on the corrupted target domain in Fig.~\ref{fig:qualitative}. As seen from the qualitative results, direct testing source domain model~(Direct Test) tends to miss small objects subject to strong corruption, e.g. the objects in the sky in row 1 and the faraway vehicles in row 3 and 4. TTA with self-training~\cite{xu2021end} and TTAC~\cite{su2022revisiting} improves detection results from the baseline by being able to identify smaller objects and better localization of spatial extent. Finally, STFAR achieves more stable detection results with more smaller objects being detected with more accurate spatial extent.

\begin{table*}[]
\caption{Test-time adaptation object detection results on  \textbf{  COCO } → \textbf{ COCO-C} dataset.}
\label{tab: COCO}
\centering
\resizebox{\textwidth}{!}{%
\begin{tabular}{c|c|ccccccccccccccc|c}
\toprule
Backbone & Methods & Brit & Contr & Defoc & Elast & Fog & Frost & Gauss & Glass & Impul & Jpeg & Motn & Pixel & Shot & Snow & Zoom & mAP \\
\midrule
\multirow{9.5}{*}{ResNet-50} & 
Clean & - & - & - & - & - & - & - & - & - &  - & - & - & - & - & - & 44.6\\
\cmidrule(lr){2-18}
& Direct Test &38.4  &22.9  &12.9  &16.5  &38.9  &24.0  &8.2  &4.7  &9.1  &13.2  &9.1  &6.2  &10.0  &19.8  &4.9  &15.9  \\
& BN~\cite{ioffe2015batch} & 15.2 & 3.4 &1.7  &7.3 &13.6  & 8.3 & 1.4 & 0.8 & 1.5 & 3.1 & 1.8 & 2.2 & 1.8 & 5.8 &2.0  & 4.7 \\
& TENT~\cite{wang2020tent} &8.5& 5.6 & 0.5 &5.0  &9.7&6.4& 1.5 & 0.5 &1.6&2.2  & 1.6 &2.4  &1.7  &5.4  & 0.8 & 3.6 \\
& T3A~\cite{iwasawa2021test} & 28.8 & 15.9 & 8.3  &  11.3& 28.9  & 17.2 &4.6  & 3.1 & 5.2 & 9.0 & 5.8 & 4.1 & 5.8 & 13.8 &3.5 & 11.0 \\
& SHOT~\cite{liang2020we} & \textbf{40.9} & 26.6 &14.7  &  19.7& \textbf{41.5} & 26.7 & 11.0 &7.2  &12.1  &16.4  & 11.0 & 9.7 &13.0  &22.0  & 6.4 & 18.6 \\
& Self-Training~\cite{xu2021end} & 38.1 & 28.4 & 14.7 & 25.5 & 38.5 & 27.9 & 16.7 & 11.4 & 18.8 & 23.8 & 16.0 & 24.5 & 18.6 & 27.6 & 7.8 & 22.6 \\
& TTAC~\cite{su2022revisiting} & 38.3 & 29.5 & 15.1 & 28.2 & 39.0 & 28.5 &16.8  &14.3  & 18.0 & 23.2 & 14.3 & 24.8 & 19.3 & 26.7 &8.7  &23.0  \\
& STFAR~(Ours) &39.1  & \textbf{31.1}  & \textbf{16.8}  & \textbf{29.0}  &39.0  & \textbf{29.2}  & \textbf{19.2}  & \textbf{15.4}  & \textbf{20.1}  & \textbf{26.1}  & \textbf{17.2}  & \textbf{28.3}  & \textbf{21.0}  & \textbf{29.5}  & \textbf{10.2}  & \textbf{24.7}  \\
\midrule
\multirow{5.5}{*}{ResNet-101}
& Clean & - & - & - & - & - & - & - & - & - &  - & - & - & - & - & - & 47.6\\
\cmidrule(lr){2-18}
&
Direct Test & 41.8 & 26.8 & 15.1 & 18.9 & 42.5 & 26.9 & 11.7 & 7.1 & 12.2 
& 16.0 & 10.9 & 8.7 & 13.8 & 23.3 & 5.5 & 18.7\\
& Self-Training~\cite{xu2021end}  & 41.6 & 31.8 & 18.4 & 28.9 & 41.5 & 31.6 &18.3 &17.1 & 22.2 & 22.8 &17.9 &27.8 &21.1 & 31.5 & 8.5& 25.4\\
& TTAC~\cite{su2022revisiting} & 42.3 & 33.5 & 18.3 & 30.7 & 42.6 & 31.5 & 21.2 & 17.7 & 22.1 &24.9  &16.9 & 26.3 & 23.1 & 30.0 & 9.7 & 26.1\\
& STFAR~(Ours)& \textbf{42.9} & \textbf{34.2} & \textbf{19.2} & \textbf{32.7} & \textbf{43.0} & \textbf{33.1} & \textbf{23.5} & \textbf{19.1} & \textbf{24.2} & \textbf{28.4} & \textbf{19.5} & \textbf{30.8} & \textbf{25.3}& \textbf{33.4} & \textbf{11.2} &\textbf{28.0} \\
\bottomrule
\end{tabular}%
}
\end{table*}

\begin{table*}[]
\caption{Test-time adaptation object detection results on \textbf{PASCAL} → \textbf{PASCAL-C} dataset with ResNet-50 as backbone.}
\label{tab:pascal}
\centering
\resizebox{0.9\linewidth}{!}{%
\begin{tabular}{c|ccccccccccccccc|c}
\toprule
Methods & Brit & Contr & Defoc & Elast & Fog & Frost & Gauss & Glass & Impul & Jpeg & Motn & Pixel & Shot & Snow & Zoom & mAP \\
\midrule
Clean & - & - & - & - & - & - & - & - & - & - & - & - & - & - & - & 80.4\\
\midrule
Direct Test &69.5  &23.8  &16.7  &42.7  &64.2  &41.7  &11.9  &13.0  &13.6  &35.8  &18.4  &26.0  &16.0  &38.2  &25.7  & 30.5 \\
BN~\cite{ioffe2015batch} & 39.5 & 20.6 & 7.4 & 17.1 & 35.3 & 22.1 & 4.7 & 4.5 &5.1  &10.5  & 9.8 & 9.1 & 6.8 & 19.1 & 13.4 & 15.0 \\
TENT~\cite{wang2020tent} & 19.6 & 9.9 &2.6  &11.0  &19.0  & 13.7 & 3.1 & 2.5 & 3.3 & 4.5 & 5.3 &8.8  & 4.0 &12.8  &4.8  & 8.3 \\
T3A~\cite{iwasawa2021test} & 36.9 & 12.5 & 11.0  &  19.7 & 32.7  &20.6 & 6.1  & 6.4 & 6.5 & 14.8 & 10.1 & 13.2 &8.4 & 16.8 &13.8 & 15.3\\
SHOT~\cite{liang2020we} &72.0& 31.7 &18.9  & 46.6 & 67.5 &45.8  &12.0  & 11.6 & 16.4 & 41.8 & 19.7 & 33.1 & 19.9 & 42.5 & 27.6 & 33.8 \\
Self-Training~\cite{xu2021end} & 67.9 & 39.3 & 2.6 & 52.5 & 65.7 & 47.2 &11.9  &20.2  & 12.1 & 29.3 &4.1  &6.9  & 17.4 & 44.9 & 9.5 & 28.8 \\
TTAC~\cite{su2022revisiting} & \textbf{72.2}  & 40.4 & 29.3 & \textbf{58.1} & \textbf{68.7} & 50.4 & 29.8 & 28.7 &33.6 & 46.4 &29.2& 46.1 & 35.1& 48.0& \textbf{34.9}  &43.4\\
STFAR~(Ours) & 67.3	 & \textbf{51.8} & \textbf{34.8} & 55.7 & 65.2	  & \textbf{50.7} & \textbf{32.4} & \textbf{34.6} & \textbf{36.3} & \textbf{49.4} & \textbf{34.6} & \textbf{55.7} & \textbf{37.8} & \textbf{50.9} &34.8   & \textbf{46.1} \\ 
\bottomrule
\end{tabular}%
}
\end{table*}

\subsection{Ablation Studies}

We carry out ablation study on  COCO-C dataset to validate the effectiveness of individual components. Specifically, we ablate self-training, global feature alignment and foreground feature alignment. We observe from the results in Tab.~\ref{tab:ablation} that self-training is effective for adapting source domain model to target distribution evidenced by the improvement from direct testing ($19.8\%\rightarrow27.6\%$). When global feature alignment is independently applied, we also observe a significant improvement from baseline ($19.8\%\rightarrow26.7\%$), though it is slightly behind self-training. As self-training is more prone to incorrect pseudo labels, we additionally utilize global feature alignment to regularize self-training, which results in an improvement from both self-training and global feature alignment alone. Finally, when foreground distribution alignment is incorporated we achieve the best performance on TTA, suggesting the effectiveness of combining self-training with distribution alignment for test-time adaptive object detection.

\subsection{Further Analysis}\label{sect:further}

In this section, we make further insights into some alternative designs of the framework, provide more qualitative results and more detailed analysis of results.

\begin{figure}[ht]
\centering
\includegraphics[width=0.85\linewidth]{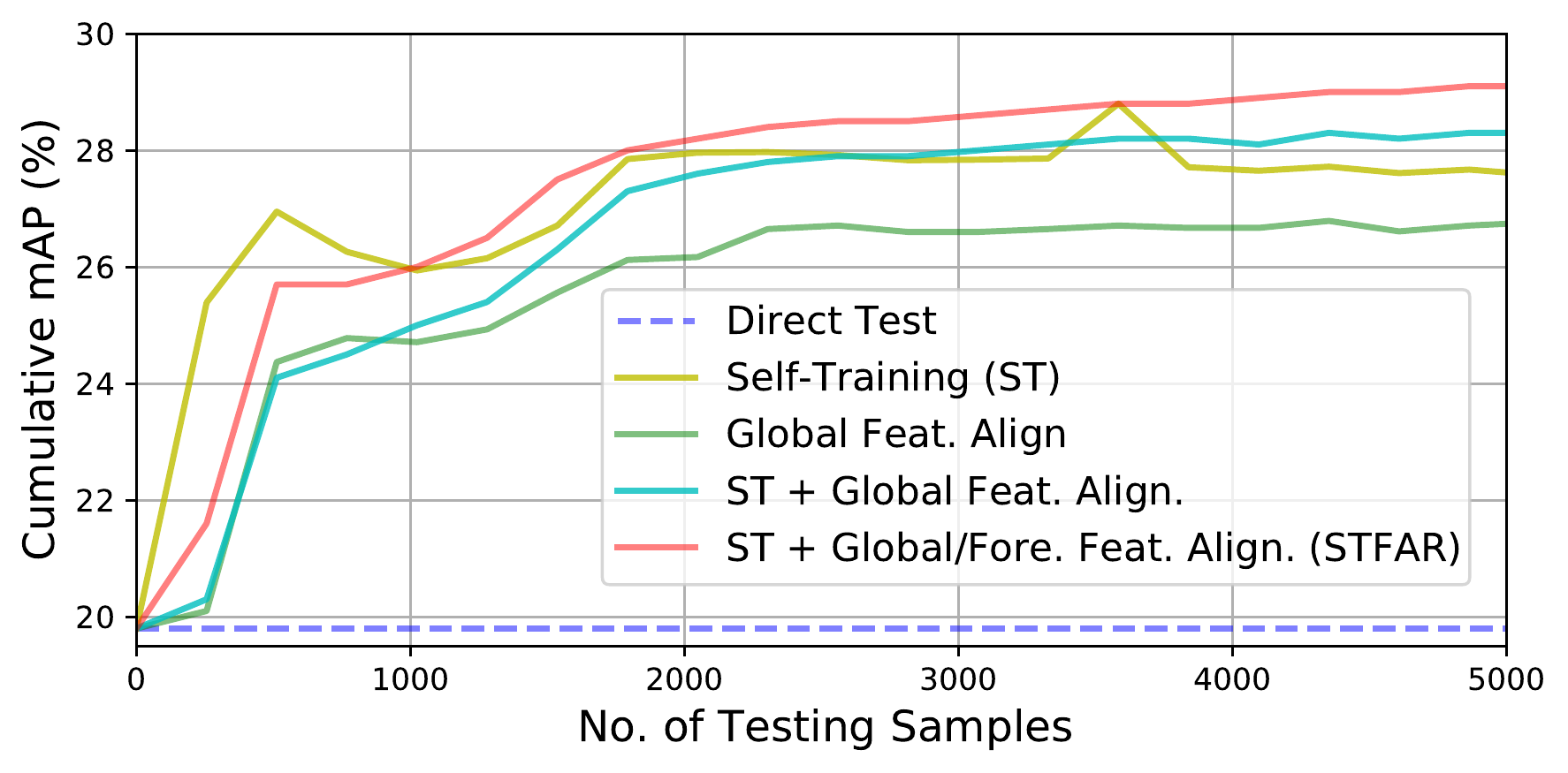}
\caption{The cumulative test-time adaptation performance (mAP) on   COCO-C dataset.}
\label{fig:cumulative}
\end{figure}

\noindent\textbf{Cumulative Performance}: We investigate the cumulative performance of ablated models on   COCO-C dataset. Good TTA methods should maintain stable or gradually increasing performance during TTA. As shown in Fig.~\ref{fig:cumulative}, we make the following observations. First, self-training~(ST) alone is a strong baseline as it picks up accuracy at very early stage of TTA. However, ST alone may suffer from the accumulation of incorrect pseudo labels and as more testing samples are seen, the performance of ST starts to decrease, partially owing to the confirmation bias. Second, global feature alignment alone is a relatively stable method for TTA, however, it struggles to further improve the performance at the late stage of TTA. Finally, STFAR benefits from the advantage of ST in fast convergence and still maintains its performance at the late stage of TTA due to the regularization of feature alignment.

\begin{table}[]
\caption{Test-time adaptation object detection results on \textbf{Cityscapes} → \textbf{Foggy-Cityscapes} dataset with ResNet-50 as backbone.}
\label{tab:cityscapes}
\centering
\resizebox{\columnwidth}{!}{%
\begin{tabular}{c|cccccccc|c}
\toprule
Methods & Pson & Rder & Car & Tuck & Bus & Train & Mcle & Bcle & mAP \\
\midrule
Clean & - & - & - & - & - & - & - & - & 40.3\\
\midrule
Direct Test     &  23.9    &26.2      &32.6     &12.2      &25.4     &4.5       &11.9      &20.9      &19.7     \\
BN~\cite{ioffe2015batch}  &  12.6 & 18.3 & 17.8 & 6.8 &0.6 & 1.0 &  2.3  &11.9 &8.9 \\
TENT~\cite{wang2020tent}  & 13.2 & 15.7 & 21.2 & 5.7 & 6.2 & 0.0 & 1.8  & 8.0 &  8.9  \\
T3A~\cite{iwasawa2021test}  &  22..6  & 23.0& 31.9 & 7.7 & 14.8 & 1.0 & 7.9& 19.7 &  16.6\\
SHOT~\cite{liang2020we}  &26.7&30.3 &36.9& \textbf{16.8}  &28.9& 6.4 & 14.3  & 23.3& 23.0\\
Self-Training~\cite{xu2021end} & 27.7  & 30.8 & 41.4 &  12.8 & 27.4 &  4.2 &  14.8  &26.1 & 23.1   \\
TTAC~\cite{su2022revisiting}  & 24.5 & 27.3 &33.4 & 14.6 & 26.1 &5.8  & 14.1 & 21.5 & 20.9 \\ 
\midrule
STFAR~(Ours) & \textbf{28.8} & \textbf{32.0} & \textbf{42.4} & 15.1 & \textbf{30.1} & \textbf{11.2} & \textbf{15.5} & \textbf{26.2} & \textbf{25.1} \\ 
\bottomrule
\end{tabular}%
}
\vspace{-0.1cm}
\end{table}

\begin{table}[]
\centering
\caption{Ablation study on \textbf{ COCO-C} dataset. The mean Average Precision (mAP) is reported on the Snow corruption on   COCO-C validation set.}
\label{tab:ablation}
\resizebox{0.95\linewidth}{!}{%
\begin{tabular}{cccc}
    \toprule
    Self-Training & Global Feat. Align. & Foreground Feat. Align. & mAP \\
    \midrule
    -     & -     & -     & 19.8 \\
    \checkmark & -     & -     & 27.6 \\
    -     & \checkmark & -     & 26.7 \\
    \checkmark & \checkmark & -     & 28.5 \\
    \checkmark & \checkmark & \checkmark & \textbf{29.5} \\
    \bottomrule
    \end{tabular}%
}
\vspace{-0.1cm}
\end{table}

\begin{table}[]
\caption{Analysis of the choice of model update parameters. The mean Average Precision (mAP) is reported on the Snow corruption on COCO-C validation set.}
\label{tab:network_weight}
\centering
\resizebox{0.85\linewidth}{!}{%
\begin{tabular}{c|c}
\toprule
Methods  & mAP \\
\midrule
Direct Test & 19.8 \\
\midrule
Update BatchNorm~(BN) statistics &  5.8\\
Update BN affine projection~(Update the stats) &  7.3\\
Update BN affine projection~(Freeze the stats) &  24.6\\
Update all network  except BN Layer  &   27.2   \\
Update  backbone except BN Layer (Ours)& \textbf{29.5}      \\ 
\bottomrule
\end{tabular}%
}
\vspace{-0.3cm}
\end{table}

\noindent\textbf{Alternative Weights Update}: By default, STFAR only updates the backbone weights during TTA as it allows reusing RPN and RCNN networks which are less likely to be affected the corruptions in target domain. In this section, we examine the alternative subsets of weights to update during TTA. As shown in Tab.~\ref{tab:network_weight}, when batchnorm statistics are allowed to be updated, we observe a significant compromise of mAP. When BN statistics are frozen and only the affine projection parameters are allowed to be updated, we observe an increase from baseline method. When all model weights, including RPN and RCNN, are updated, the performance is still inferior to STFAR which only updates the backbone weights. To conclude, updating only the backbone weights during TTA is more effective than alternative updating strategies.

\noindent\textbf{Alternative Backbone}: We further implemented test-time adaptive object dtection on COCO-C with ResNet-101 as backbone network with results presented in Tab.~\ref{tab: COCO}. Again, we observe consistently improvement of STFAR over TTAC or Self-Training alone, suggesting the effectiveness of the proposed method.



\section{Conclusion}

In this work, we investigate a realistic setting for adapting source domain model to target distribution data subject to natural corruptions. Testing data distribution is assumed to be unknown before the inference stage. Model weights are then adapted to target domain at test-time by combining self-training with feature distribution alignment regularization. Three standard object detection datasets are converted for the evaluation of test-time adaptive object detection task. Extensive comparisons with test-time adaptation methods confirm the effectiveness of the proposed method.




{\small
\bibliographystyle{ieee_fullname}
\bibliography{egbib}

\begin{thebibliography}{10}\itemsep=-1pt

\bibitem{arazo2020pseudo}
Eric Arazo, Diego Ortego, Paul Albert, Noel~E O’Connor, and Kevin McGuinness.
\newblock Pseudo-labeling and confirmation bias in deep semi-supervised
  learning.
\newblock In {\em International Joint Conference on Neural Networks}, 2020.

\bibitem{carion2020end}
Nicolas Carion, Francisco Massa, Gabriel Synnaeve, Nicolas Usunier, Alexander
  Kirillov, and Sergey Zagoruyko.
\newblock End-to-end object detection with transformers.
\newblock In {\em European Conference on Computer Vision}, 2020.

\bibitem{Chen_2020_CVPR}
Chaoqi Chen, Zebiao Zheng, Xinghao Ding, Yue Huang, and Qi Dou.
\newblock Harmonizing transferability and discriminability for adapting object
  detectors.
\newblock In {\em Proceedings of the IEEE/CVF Conference on Computer Vision and
  Pattern Recognition}, 2020.

\bibitem{chen2022contrastive}
Dian Chen, Dequan Wang, Trevor Darrell, and Sayna Ebrahimi.
\newblock Contrastive test-time adaptation.
\newblock In {\em Proceedings of the IEEE/CVF Conference on Computer Vision and
  Pattern Recognition}, 2022.

\bibitem{chen2018}
Y. Chen, W. Li, C. Sakaridis, D. Dai, and L.~Van Gool.
\newblock Domain adaptive faster r-cnn for object detection in the wild.
\newblock In {\em IEEE/CVF Conference on Computer Vision and Pattern
  Recognition}, 2018.

\bibitem{Chu2023adver}
Qiaosong Chu, Shuyan Li, Guangyi Chen, Kai Li, and Xiu Li.
\newblock Adversarial alignment for source free object detection, 2023.

\bibitem{cordts2016cityscapes}
Marius Cordts, Mohamed Omran, Sebastian Ramos, Timo Rehfeld, Markus Enzweiler,
  Rodrigo Benenson, Uwe Franke, Stefan Roth, and Bernt Schiele.
\newblock The cityscapes dataset for semantic urban scene understanding.
\newblock In {\em Proceedings of the IEEE conference on computer vision and
  pattern recognition}, 2016.

\bibitem{everingham2015pascal}
Mark Everingham, SM~Ali Eslami, Luc Van~Gool, Christopher~KI Williams, John
  Winn, and Andrew Zisserman.
\newblock The pascal visual object classes challenge: A retrospective.
\newblock {\em International Journal of Computer Vision}, 2015.

\bibitem{ganin2015unsupervised}
Yaroslav Ganin and Victor Lempitsky.
\newblock Unsupervised domain adaptation by backpropagation.
\newblock In {\em International Conference on Machine Learning}, 2015.

\bibitem{girshick2015fast}
Ross Girshick.
\newblock Fast r-cnn.
\newblock In {\em Proceedings of the IEEE International Conference on Computer
  Vision}, 2015.

\bibitem{gongnote2022}
Taesik Gong, Jongheon Jeong, Taewon Kim, Yewon Kim, Jinwoo Shin, and Sung-Ju
  Lee.
\newblock Note: Robust continual test-time adaptation against temporal
  correlation.
\newblock In {\em Advances in Neural Information Processing Systems}, 2022.

\bibitem{he2016deep}
Kaiming He, Xiangyu Zhang, Shaoqing Ren, and Jian Sun.
\newblock Deep residual learning for image recognition.
\newblock In {\em Proceedings of the IEEE conference on computer vision and
  pattern recognition}, pages 770--778, 2016.

\bibitem{He_2022_CVPR}
Mengzhe He, Yali Wang, Jiaxi Wu, Yiru Wang, Hanqing Li, Bo Li, Weihao Gan, Wei
  Wu, and Yu Qiao.
\newblock Cross domain object detection by target-perceived dual branch
  distillation.
\newblock In {\em Proceedings of the IEEE/CVF Conference on Computer Vision and
  Pattern Recognition (CVPR)}, pages 9570--9580, June 2022.

\bibitem{He_2019_ICCV}
Zhenwei He and Lei Zhang.
\newblock Multi-adversarial faster-rcnn for unrestricted object detection.
\newblock In {\em Proceedings of the IEEE/CVF International Conference on
  Computer Vision}, 2019.

\bibitem{NEURIPS2021_1dba5eed}
Jiaxing Huang, Dayan Guan, Aoran Xiao, and Shijian Lu.
\newblock Model adaptation: Historical contrastive learning for unsupervised
  domain adaptation without source data.
\newblock In M. Ranzato, A. Beygelzimer, Y. Dauphin, P.S. Liang, and J.~Wortman
  Vaughan, editors, {\em Advances in Neural Information Processing Systems},
  volume~34, pages 3635--3649. Curran Associates, Inc., 2021.

\bibitem{ioffe2015batch}
Sergey Ioffe and Christian Szegedy.
\newblock Batch normalization: Accelerating deep network training by reducing
  internal covariate shift.
\newblock In {\em International conference on machine learning}, pages
  448--456. pmlr, 2015.

\bibitem{iwasawa2021test}
Yusuke Iwasawa and Yutaka Matsuo.
\newblock Test-time classifier adjustment module for model-agnostic domain
  generalization.
\newblock {\em Advances in Neural Information Processing Systems},
  34:2427--2440, 2021.

\bibitem{Khodabandeh_2019_ICCV}
Mehran Khodabandeh, Arash Vahdat, Mani Ranjbar, and William~G. Macready.
\newblock A robust learning approach to domain adaptive object detection.
\newblock In {\em Proceedings of the IEEE/CVF International Conference on
  Computer Vision}, 2019.

\bibitem{Kim_2019_ICCV}
Seunghyeon Kim, Jaehoon Choi, Taekyung Kim, and Changick Kim.
\newblock Self-training and adversarial background regularization for
  unsupervised domain adaptive one-stage object detection.
\newblock In {\em Proceedings of the IEEE/CVF International Conference on
  Computer Vision (ICCV)}, October 2019.

\bibitem{Kim_2019_CVPR}
Taekyung Kim, Minki Jeong, Seunghyeon Kim, Seokeon Choi, and Changick Kim.
\newblock Diversify and match: A domain adaptive representation learning
  paradigm for object detection.
\newblock In {\em Proceedings of the IEEE/CVF Conference on Computer Vision and
  Pattern Recognition}, 2019.

\bibitem{li2020model}
Rui Li, Qianfen Jiao, Wenming Cao, Hau-San Wong, and Si Wu.
\newblock Model adaptation: Unsupervised domain adaptation without source data.
\newblock In {\em Proceedings of the IEEE/CVF conference on computer vision and
  pattern recognition}, 2020.

\bibitem{li2021}
Shuai Li, Jianqiang Huang, Xian-Sheng Hua, and Lei Zhang.
\newblock Category dictionary guided unsupervised domain adaptation for object
  detection.
\newblock In {\em Proceedings of the AAAI Conference on Artificial
  Intelligence}, 2021.

\bibitem{li2022source}
Shuaifeng Li, Mao Ye, Xiatian Zhu, Lihua Zhou, and Lin Xiong.
\newblock Source-free object detection by learning to overlook domain style.
\newblock In {\em Proceedings of the IEEE/CVF Conference on Computer Vision and
  Pattern Recognition}, 2022.

\bibitem{li2021free}
Xianfeng Li, Weijie Chen, Di Xie, Shicai Yang, Peng Yuan, Shiliang Pu, and
  Yueting Zhuang.
\newblock A free lunch for unsupervised domain adaptive object detection
  without source data.
\newblock In {\em Proceedings of the AAAI Conference on Artificial
  Intelligence}, 2021.

\bibitem{liang2020we}
Jian Liang, Dapeng Hu, and Jiashi Feng.
\newblock Do we really need to access the source data? source hypothesis
  transfer for unsupervised domain adaptation.
\newblock In {\em International Conference on Machine Learning}, 2020.

\bibitem{lim2023ttn}
Hyesu Lim, Byeonggeun Kim, Jaegul Choo, and Sungha Choi.
\newblock {TTN}: A domain-shift aware batch normalization in test-time
  adaptation.
\newblock In {\em International Conference on Learning Representations}, 2023.

\bibitem{Lin_2017_ICCV}
Tsung-Yi Lin, Priya Goyal, Ross Girshick, Kaiming He, and Piotr Dollar.
\newblock Focal loss for dense object detection.
\newblock In {\em Proceedings of the IEEE International Conference on Computer
  Vision (ICCV)}, Oct 2017.

\bibitem{lin2014microsoft}
Tsung-Yi Lin, Michael Maire, Serge Belongie, James Hays, Pietro Perona, Deva
  Ramanan, Piotr Doll{\'a}r, and C~Lawrence Zitnick.
\newblock Microsoft coco: Common objects in context.
\newblock In {\em European Conference on Computer Vision}, 2014.

\bibitem{liu2021cycle}
Hong Liu, Jianmin Wang, and Mingsheng Long.
\newblock Cycle self-training for domain adaptation.
\newblock In {\em Advances in Neural Information Processing Systems}, 2021.

\bibitem{liu2021ttt++}
Yuejiang Liu, Parth Kothari, Bastien Van~Delft, Baptiste Bellot-Gurlet, Taylor
  Mordan, and Alexandre Alahi.
\newblock Ttt++: When does self-supervised test-time training fail or thrive?
\newblock In {\em Advances in Neural Information Processing Systems}, 2021.

\bibitem{michaelis2019dragon}
Claudio Michaelis, Benjamin Mitzkus, Robert Geirhos, Evgenia Rusak, Oliver
  Bringmann, Alexander~S. Ecker, Matthias Bethge, and Wieland Brendel.
\newblock Benchmarking robustness in object detection: Autonomous driving when
  winter is coming.
\newblock {\em arXiv preprint arXiv:1907.07484}, 2019.

\bibitem{conditionalGAN}
Mehdi Mirza and Simon Osindero.
\newblock Conditional generative adversarial nets, 2014.

\bibitem{pmlr-v162-niu22a}
Shuaicheng Niu, Jiaxiang Wu, Yifan Zhang, Yaofo Chen, Shijian Zheng, Peilin
  Zhao, and Mingkui Tan.
\newblock Efficient test-time model adaptation without forgetting.
\newblock In Kamalika Chaudhuri, Stefanie Jegelka, Le Song, Csaba Szepesvari,
  Gang Niu, and Sivan Sabato, editors, {\em Proceedings of the 39th
  International Conference on Machine Learning}, volume 162 of {\em Proceedings
  of Machine Learning Research}, pages 16888--16905. PMLR, 17--23 Jul 2022.

\bibitem{niu2023towards}
Shuaicheng Niu, Jiaxiang Wu, Yifan Zhang, Zhiquan Wen, Yaofo Chen, Peilin Zhao,
  and Mingkui Tan.
\newblock Towards stable test-time adaptation in dynamic wild world.
\newblock In {\em The Eleventh International Conference on Learning
  Representations}, 2023.

\bibitem{ren2015faster}
Shaoqing Ren, Kaiming He, Ross Girshick, and Jian Sun.
\newblock Faster r-cnn: Towards real-time object detection with region proposal
  networks.
\newblock {\em IEEE Transactions on Pattern Analysis and Machine Intelligence},
  2016.

\bibitem{Saito_2019_CVPR}
Kuniaki Saito, Yoshitaka Ushiku, Tatsuya Harada, and Kate Saenko.
\newblock Strong-weak distribution alignment for adaptive object detection.
\newblock In {\em Proceedings of the IEEE/CVF Conference on Computer Vision and
  Pattern Recognition}, 2019.

\bibitem{sakaridis2018semantic}
Christos Sakaridis, Dengxin Dai, and Luc Van~Gool.
\newblock Semantic foggy scene understanding with synthetic data.
\newblock {\em International Journal of Computer Vision}, 2018.

\bibitem{sohn2020fixmatch}
Kihyuk Sohn, David Berthelot, Nicholas Carlini, Zizhao Zhang, Han Zhang,
  Colin~A Raffel, Ekin~Dogus Cubuk, Alexey Kurakin, and Chun-Liang Li.
\newblock Fixmatch: Simplifying semi-supervised learning with consistency and
  confidence.
\newblock In {\em Advances in neural information processing systems}, 2020.

\bibitem{su2022revisiting}
Yongyi Su, Xun Xu, and Kui Jia.
\newblock Revisiting realistic test-time training: Sequential inference and
  adaptation by anchored clustering.
\newblock In {\em Advances in Neural Information Processing Systems}, 2022.

\bibitem{su2023revisiting}
Yongyi Su, Xun Xu, Tianrui Li, and Kui Jia.
\newblock Revisiting realistic test-time training: Sequential inference and
  adaptation by anchored clustering regularized self-training.
\newblock {\em arXiv preprint arXiv:2303.10856}, 2023.

\bibitem{sun2020test}
Yu Sun, Xiaolong Wang, Zhuang Liu, John Miller, Alexei Efros, and Moritz Hardt.
\newblock Test-time training with self-supervision for generalization under
  distribution shifts.
\newblock In {\em International conference on machine learning}, 2020.

\bibitem{NIPS2017_68053af2}
Antti Tarvainen and Harri Valpola.
\newblock Mean teachers are better role models: Weight-averaged consistency
  targets improve semi-supervised deep learning results.
\newblock In I. Guyon, U.~Von Luxburg, S. Bengio, H. Wallach, R. Fergus, S.
  Vishwanathan, and R. Garnett, editors, {\em Advances in Neural Information
  Processing Systems}, 2017.

\bibitem{tarvainen2017mean}
Antti Tarvainen and Harri Valpola.
\newblock Mean teachers are better role models: Weight-averaged consistency
  targets improve semi-supervised deep learning results.
\newblock {\em Advances in neural information processing systems}, 2017.

\bibitem{wang2020tent}
Dequan Wang, Evan Shelhamer, Shaoteng Liu, Bruno Olshausen, and Trevor Darrell.
\newblock Tent: Fully test-time adaptation by entropy minimization.
\newblock In {\em International Conference on Learning Representations}, 2020.

\bibitem{wang_2021_tip}
Hongsong Wang, Shengcai Liao, and Ling Shao.
\newblock Afan: Augmented feature alignment network for cross-domain object
  detection.
\newblock {\em IEEE Transactions on Image Processing}, 2021.

\bibitem{wang2022continual}
Qin Wang, Olga Fink, Luc Van~Gool, and Dengxin Dai.
\newblock Continual test-time domain adaptation.
\newblock In {\em Proceedings of the IEEE/CVF Conference on Computer Vision and
  Pattern Recognition}, pages 7201--7211, 2022.

\bibitem{Xie_2019_ICCV}
Rongchang Xie, Fei Yu, Jiachao Wang, Yizhou Wang, and Li Zhang.
\newblock Multi-level domain adaptive learning for cross-domain detection.
\newblock In {\em Proceedings of the IEEE/CVF International Conference on
  Computer Vision (ICCV) Workshops}, Oct 2019.

\bibitem{Xu_2020_CVPR}
Chang-Dong Xu, Xing-Ran Zhao, Xin Jin, and Xiu-Shen Wei.
\newblock Exploring categorical regularization for domain adaptive object
  detection.
\newblock In {\em Proceedings of the IEEE/CVF Conference on Computer Vision and
  Pattern Recognition}, 2020.

\bibitem{xu2021end}
Mengde Xu, Zheng Zhang, Han Hu, Jianfeng Wang, Lijuan Wang, Fangyun Wei, Xiang
  Bai, and Zicheng Liu.
\newblock End-to-end semi-supervised object detection with soft teacher.
\newblock In {\em Proceedings of the IEEE/CVF International Conference on
  Computer Vision}, 2021.

\bibitem{Zhao2020}
Zhen Zhao, Yuhong Guo, and Jieping Ye.
\newblock Bi-dimensional feature alignment for cross-domain object detection.
\newblock In {\em Computer Vision -- ECCV 2020 Workshops}, pages 671--686,
  2020.

\end{thebibliography}
}

In this supplementary material, we first discuss the limitations of the proposed TTA method, STFAR. Then, we further carry out additional evaluations on stability of TTA algorithm, alternative designs for feature alignment, compatibility with alternative transformer based detection framework, etc.

\section{Appendix}

\subsection{Limitations}

We discuss the limitations of our proposed method from three perspectives. First, in order to regularize  self-training, we perform feature alignment between source and target domains, which needs us to acquire source domain distribution information. This can achieved by running inference on source domain data. Second, the continual test-time adaptation~\cite{wang2022continual} is a valuable TTA classification scenario, yet the proposed setting named TTAOD in this paper is a more common and realistic Test-time detection scenario and we expect to conduct research on the more challenging continual test-time adaptive object detection task in the future. Third, we reveal that multiple updating steps are performed on a mini batch for better convergence. This would introduce more computation overhead and may pose challenges to the demand for close to real-time test-time adaptation. As shown in Tab.~\ref{tab:cityscapes}, we carry out additional evaluations on different numbers of update steps, and the observations suggest that updating 3 times achieves the best result.

\begin{table}[htbp]
    \centering
    \caption{Performance Analysis of Random Seed Ablation Study on snow corruption of COCO-C dataset. Each number indicates the mean Average Precision~(mAP).}
    \resizebox{\columnwidth}{!}{%
        \begin{tabular}{cc|ccccc|c}
        \toprule
           \multirow{2.5}{*}{Clean} & \multirow{2.5}{*}{Direct Test} & \multicolumn{6}{c}{STFAR with different seeds}\\
        \cmidrule{3-8}
            & & 0 & 1 & 42 & 789 & 2000 & Avg$\pm$Std\\
        \midrule
            44.6 & 19.8 & 29.6 & 29.6 & 29.3 & 29.3 & 29.7 & 29.5$\pm$0.2\\
        \bottomrule
        \end{tabular}
    }
    \label{tab:randseed}
\end{table}

\subsection{Stability of TTA Algorithm}

We investigate the stability of STFAR algorithm by running multiple times with different random seeds.
Tab.~\ref{tab:randseed} presents the mean Average Precision (mAP) over five different seeds evaluated on the Snow corruption on   COCO-C validation set. All the seeds produced similar results, with the mAP values ranging from 29.3 to 29.7. The results suggest that the model is relatively stable and robust to different random seeds, as the performance does not vary significantly between the different runs.

\subsection{Alternative Designs for Global Feature Alignment}
\begin{table}[]
\caption{Alternative choices of features for global feature alignment. The mean Average Precision (mAP) is reported on the Snow corruption of COCO-C dataset.}
\label{tab:fpn_network}
\centering
\resizebox{0.5\linewidth}{!}{%
\begin{tabular}{c|c}
\toprule
Method  & mAP \\
\midrule
Clean &  44.6\\
\midrule
Direct Test & 19.8 \\
\midrule
align FPN layer 0 &29.3 \\
alignt FPN layer 1 &  29.4\\
alignt FPN layer 2 &  29.3\\
align FPN layer 3 &  29.4 \\
align FPN layer 4 (Ours)& \textbf{29.5}      \\ 
\midrule
align ResNet50 layer & 28.5      \\ 
\bottomrule
\end{tabular}%
}
\end{table}

To identify the most suitable global feature for object detection, we select features across different layers for global feature alignment (see Table~\ref{tab:fpn_network}). Specifically, we utilized the Faster Rcnn architecture with Feature Pyramid Networks (FPN) and aligned features from different FPN layers. Our results showed that feature alignment across different layers yieldes similar performance, indicating that features from multiple layers can be used for object detection. By aligning the global features extracted from the FPN layers, we can ensure that the network focus on the most informative and discriminative features for object detection tasks.

Furthermore, we also conducted feature alignment using features from Resnet50 (without FPN features) and found that although there was some improvement compared to direct testing, however, the results are still inferior to those obtained using FPN layer features. One possible reason for this is that FPN has a pyramid-like structure, which enables it to capture features at multiple scales.
In contrast, Resnet50 only has a single-scale feature extractor, which may limit its ability to capture features at different scales. Therefore, selecting FPN features for global feature alignment can improve the performance of the network in object detection tasks. Overall, our findings suggest that using features from a multi-scale feature extractor like FPN may lead to better performance than using features from a single-scale feature extractor like Resnet50 for object detection tasks. These findings provide insights into the optimal selection of global features for object detection tasks.

\subsubsection{Alternative Designs for Foreground Feature Alignment}
\begin{table}[]
\caption{Alternative designs for the foreground feature alignment. }
\label{tab:proposal_align}
\centering
\resizebox{0.8\linewidth}{!}{%
\begin{tabular}{c|c}
\toprule
Method  & mAP \\
\midrule
Clean &  44.6\\
\midrule
Direct Test & 19.8 \\
\midrule
only align foreground feature  &23.5 \\
only align foreground class-wise feature &22.1 \\
\bottomrule
\end{tabular}%
}
\end{table}

In addition to global feature alignment at the FPN level, we also investigated feature alignment at the foreground level. Specifically, we compared foreground alignment with and without class information in Table~\ref{tab:proposal_align}. Our experiments on the COCO-C dataset shows that foreground alignment without class information achieves better results. \par

We attribute the observation to the reason that the non-category aligned features contain more domain-invariant information, which could help reduce domain shift and improve domain adaptation. Since object detection is a more complex task than classification, with additional challenges such as varying object scales, poses, and occlusions, the non-category aligned foreground features may be more robust to such variations and better suit for detecting objects in the target domain. Moreover, during Test-time Adaptation, the class labels are obtained for the proposals through pseudo labeling, which may contain noisy labels errors, thus prohibiting good estimations of foreground feature distributions.
Therefore, in object detection tasks, using foreground alignment without class information may be more suitable. 


\subsection{Compatibility with Transformer Detector}

DETR (DEtection TRansformer)~\cite{carion2020end} is a recent object detection framework that achieves state-of-the-art performance on various benchmark datasets. Unlike traditional object detection approaches that rely on region proposal mechanisms or anchor-based methods, DETR directly predicts object bounding boxes and class probabilities from a set of learned queries and feature maps using a transformer-based architecture.

We chose to apply our STFAR method to the DETR model, a one-stage object detector. Specifically, we use the DETR model provided in the mmdetection library,  consisting of two main components: an image encoder  and an object detection decoder.The released DETR model in mmdetection is  with ResNet-50 as the backbone and achieves 40.1 bounding box average precision (box AP) on clean COCO validation dataset . In our experiments, we use the global features obtained from ResNet50 as well as the foreground features obtained from the decoder for distribution alignment.

To prevent the self-training task from being biased by incorrect pseudo labels, we filtered the decoder's output boxes further using a threshold of 0.5. The results of our experiments show that our proposed method is still able to achieve good results, as demonstrated in Table ~\ref{tab:detr}.
Due to the limitations of the DETR architecture, we are not able to apply the strong self-training method~\cite{xu2021end} in DETR. Instead, we use the teacher-student framework to provide pseudo-labels for self-training and add a filtering mechanism to reduce the impact of incorrect pseudo-labels. STFAR method achieves good performance with DETR model, and the model's performance is further improved with global feature alignment or foreground feature alignment. Therefore, this method has practicality and effectiveness in object detection tasks.

\begin{table}[]
\centering
\caption{The performance of STFAR using DETR backbone on snow corruption of COCO-C dataset.}
\label{tab:detr}
\resizebox{0.95\linewidth}{!}{%
\begin{tabular}{cccc}
    \toprule
    Self-Training & Global Feat. Align. & Foreground Feat. Align. & mAP \\
    \midrule
    -     & -     & -     & 12. 1\\
    \checkmark & -     & -     &14.2  \\
    -     & \checkmark & -     & 19.6 \\
     -     & -   & \checkmark     & 16.8 \\
     \checkmark & \checkmark & -     & 20.8 \\
      \checkmark & - & \checkmark     & 17.1 \\
    \checkmark & \checkmark & \checkmark & \textbf{21.7} \\
    \bottomrule
    \end{tabular}%
}

\end{table}

\subsection{Multiple Update Steps}

\begin{table}[]
\caption{The results of different epoch times on experimental  on \textbf{Cityscapes} → \textbf{Foggy-Cityscapes} dataset with ResNet-50 as backbone.}
\label{tab:cityscapes}
\centering
\resizebox{\columnwidth}{!}{%
\begin{tabular}{c|cccccccc|c}
\toprule
\ Steps & Pson & Rder & Car & Tuck & Bus & Train & Mcle & Bcle & mAP \\
\midrule
Clean & - & - & - & - & - & - & - & - & 40.3\\
\midrule
Direct Test     &  23.9    &26.2      &32.6     &12.2      &25.4     &4.5       &11.9      &20.9      &19.7     \\
1 & 28.2& 31.8&40.1&10.5& 18.5& 2.8&12.5&23.2&21.0 \\
2 &28.7 &31.7& 41.4& 15.4& 24.2&7.4& 15.0&25.5&23.7  \\
\textbf{3}  & \textbf{28.8} & \textbf{32.0} & \textbf{42.4} & 15.1 & 30.1 & 11.2 & 15.5 & \textbf{26.2} & \textbf{25.1}\\
4 &27.8& 31.2& 41.8& 15.6& 29.6&  \textbf{11.5} & \textbf{16.8} &25.8& 25.0\\
5 & 27.4&30.5& 41.6&  \textbf{16.6} &  \textbf{32.2}& 10.4&15.9&25.8& \textbf{25.1}   \\
7 & 26.2 &28.6& 41.0&16.4&30.7&9.2&15.8&25.3&24.2 \\ 
\bottomrule
\end{tabular}%
}
\vspace{-0.1cm}
\end{table}

During test time training, data is sequentially fed into the model. To ensure model weights are sufficiently updated, we propose to apply multiple steps of gradient descent on the same batch of testing data, where steps refer to the number of times the data is used for gradient descent within each batch. In our experiments reported in the paper, we set epoch times to 2 for the COCO-C dataset (ResNet50), 3 for the FoggyCity dataset, and 1 for both the COCO-C dataset (ResNet101) and Pascal-C dataset. We further investigate the impact of different number of update steps on the model's performance in Table ~\ref{tab:cityscapes}  and find that, as epoch times increase, the model's performance gradually improves as the data is better utilized. However, too many update steps may lead to overfitting, causing a decrease in performance. Moreover, increasing update steps results in longer training time and higher consumption of computing resources, thus the number of update steps can be seen as a hyper-parameter to achieve balance between TTA
 performance and computation efficiency.

\subsection{Additional Qualitative Results}

We present qualitative results obtained by various Test-Time Training methods on the target domain in Fig.~\ref{fig:qualitative}. From the figure, it can be observed that while self-training and TTAC show some degree of improvement in detection results, they are susceptible to introducing new errors, such as duplicated or missing detections, and may not always improve the original prediction results. The proposed STFAR method, which leverages foreground and global feature alignment to regularize self-supervised training, yields more accurate and reliable results . The demonstrated effectiveness of our approach in reducing errors caused by individual methods and achieving superior overall performance suggests its potential for practical applications in real-world scenarios.

\begin{figure*}[!ht]
\centering
\includegraphics[width=0.9\textwidth]{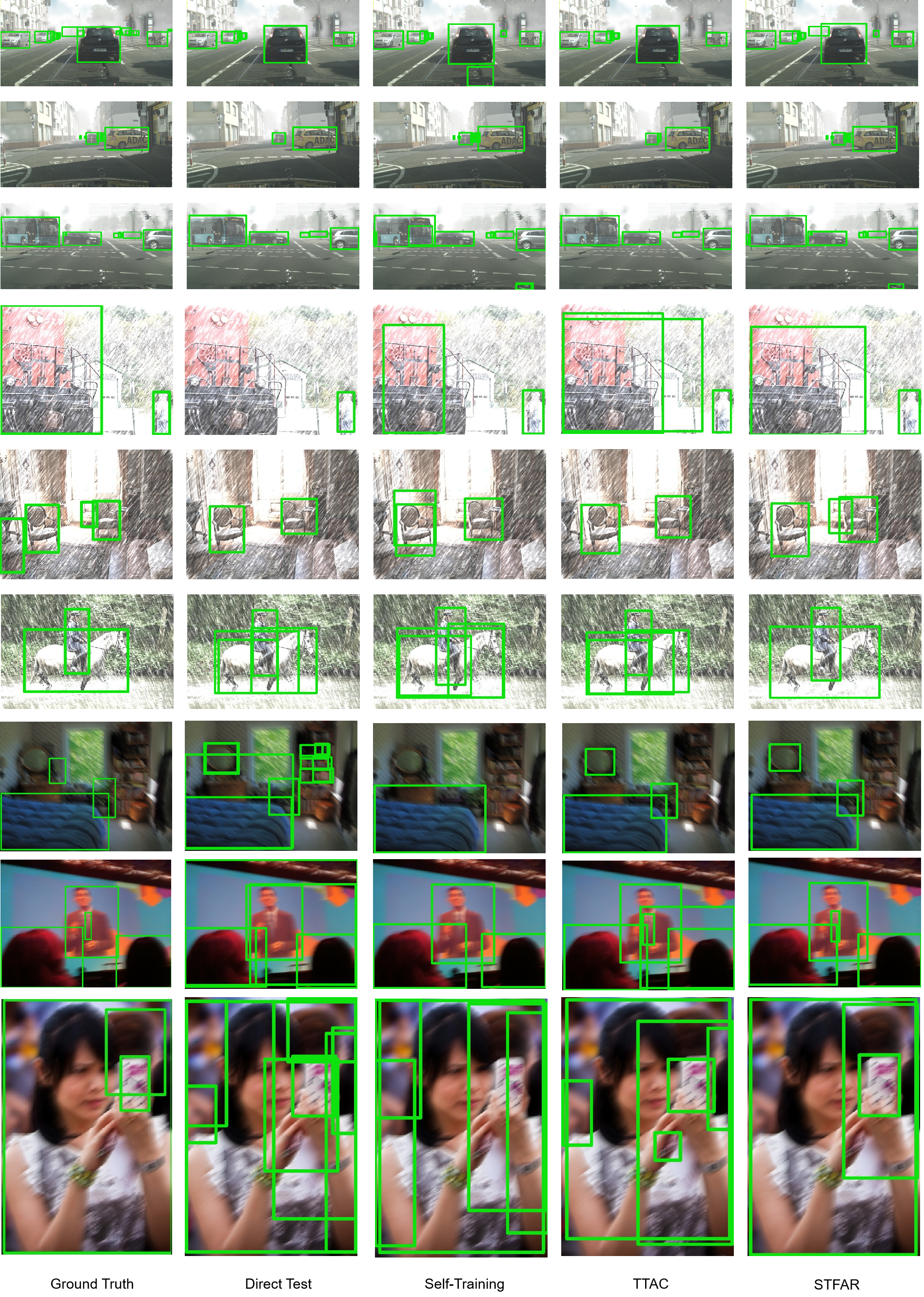}
    \caption{Illustration of detection results on the target domains. The first three rows show the scenarios of \textbf{Cityscapes} → \textbf{Foggy-Cityscapes}, while the middle three rows show \textbf{PASCAL} → \textbf{PASCAL-C}  (Snow corruption) and the last three rows  represent  \textbf{ COCO} → \textbf{ COCO-C} (Motion blur corruption). }.   
\label{fig:qualitative}
\vspace{-0.4cm}
\end{figure*}

\end{document}